\documentclass{article} 
\usepackage[T1]{fontenc}
\usepackage[final]{colm2026_conference}

\usepackage{microtype}
\usepackage{hyperref}
\usepackage{url}
\usepackage{booktabs}
\usepackage{wrapfig}
\usepackage{amsmath}
\usepackage{amssymb}
\usepackage{mathtools}
\usepackage{amsthm}
\usepackage{makecell}
\usepackage{graphicx}
\usepackage{subcaption}
\usepackage{xcolor}
\usepackage{multirow}
\usepackage{multicol}
\usepackage{pifont}
\usepackage[misc]{ifsym}

\usepackage{tcolorbox}
\tcbuselibrary{listings,skins,breakable}

\definecolor{promptbg}{RGB}{248,248,248}
\definecolor{promptframe}{RGB}{200,200,200}
\definecolor{basecolor}{RGB}{174,70,55}
\definecolor{ourscolor}{RGB}{35,105,75}

\lstdefinestyle{prompt}{
    basicstyle=\ttfamily\footnotesize,
    breaklines=true,
    columns=fullflexible,
    keepspaces=true,
    showstringspaces=false,
    frame=single,
    rulecolor=\color{gray!35},
    backgroundcolor=\color{promptbg},
    moredelim=[is][\color{basecolor}\bfseries]{<BASE>}{</BASE>},
    moredelim=[is][\color{ourscolor}\bfseries]{<OURS>}{</OURS>},
    moredelim=[is][\color{red}\bfseries]{<ERROR>}{</ERROR>}
}

\newtcblisting{promptbox}[1][]{
  listing only,
  colback=promptbg,
  colframe=promptframe,
  boxrule=0.4pt,
  arc=2pt,
  left=6pt,right=6pt,top=4pt,bottom=4pt,
  listing options={
    basicstyle=\ttfamily\footnotesize,
    breaklines=true,
    breakatwhitespace=false,
    columns=fullflexible,
    keepspaces=true,
    breakautoindent=false,
    breakindent=0pt
  },
  #1
}

\usepackage{color}
\usepackage[table]{xcolor}
\definecolor{mygreen}{RGB}{0,120,90}

\usepackage[capitalize,noabbrev]{cleveref}

\definecolor{darkblue}{rgb}{0, 0, 0.5}
\hypersetup{colorlinks=true, citecolor=mid, linkcolor=primary, urlcolor=mid}

\renewcommand{\cite}[1]{\citep{#1}}

\title{From Procedures to Permissions: Learning Precise Subagent Authorization from Agent Skills}

\author{
\mbox{
Wenkai Yu$^{2,1}$, Ning Ding$^{1}$, Wentao Zhang$^{2{~\textrm{\Letter}}}$ and Yehui Tang$^{1{~\textrm{\Letter}}}$
}\\
$^1$ Samsung Research, Beijing, China \quad\quad $^2$ Peking University\\
\texttt{ywk2022}@stu.pku.edu.cn, \texttt{ning.ding}@samsung.com \\
\texttt{wentao.zhang}@pku.edu.cn, \texttt{yehui.tang}@samsung.com \\
$^\textrm{\Letter}$~Corresponding Author
}

\begin{document}

\maketitle
\begin{abstract}
As language agents tackle increasingly complex tasks, they increasingly rely on multiple agents to coordinate specialized work, requiring decisions not only about what each subagent should do but also which tools and workspace resources it should access. Existing agent systems provide mechanisms for controlling tools and execution environments, but do not directly specify the resources required by each delegated subtask. Moreover, task outcomes and execution trajectories alone provide limited supervision for determining whether a subagent has received appropriate permissions. We observe that Agent Skills encode procedural knowledge describing the operations, tools, and files required at each step, providing a natural source of supervision for authorization. Building on this observation, we synthesize tasks and step-by-step procedures from Skills, execute them through subagent delegation in interactive environments, and use verified trajectories to construct training data for task execution and authorization. On ClawArena-Team, our method improves tool permission precision by 13.4 percentage points and workspace permission precision by 24.1 percentage points, while increasing overall task success by 11.6 percentage points for Qwen3.5-35B-A3B. On Terminal-Bench 2.1 and SkillsBench, our method remains competitive with existing agent data synthesis approaches. These results show that the procedural knowledge embedded in Agent Skills provides a source of supervision for precise subagent orchestration that task outcomes alone cannot provide.
\end{abstract}

\section{Introduction}
Large language model (LLM) agents are increasingly evolving from single-agent task solvers into systems that decompose complex tasks, use external tools, and delegate subtasks to specialized agents \citep{yao2023react,wu2023autogen,wang2024openhands}. Delegation, however, introduces a problem beyond deciding what a subagent should do: the main agent must also determine what the subagent should be allowed to access. In particular, it must decide which tools and workspace resources are necessary for the delegated subtask and which are not. Thus, effective subagent orchestration requires not only task decomposition, but also task-conditioned authorization.

This authorization decision matters because task completion alone does not guarantee an appropriate permission boundary. A subagent may successfully complete a task with access to the entire workspace and tools, even though it only needs one input file and one execution tool. Conversely, overly restrictive permissions can prevent an otherwise solvable task from being completed. Recent work demonstrates that this is a practical challenge rather than merely a theoretical concern. MiniScope \citep{zhu2025miniscope} develops least-privilege enforcement for tool-calling agents, while SkillScope \citep{wu2026skillscope} studies specific privilege violations in Agent Skills. AuthBench \citep{yan2026authbench} finds that coding agents frequently both omit necessary permissions and grant unnecessary ones, while ClawArena-Team \citep{xiong2026clawarenateam} identifies subagent management and resource allocation as important challenges in multi-agent systems. These results indicate that precise authorization is an important capability for agent orchestration, rather than merely an implementation detail.

Yet an important question remains: where should supervision for precise authorization come from? Existing agent training and data synthesis methods typically rely on task outcomes, demonstrations, or execution trajectories. Such signals indicate whether a task succeeds, but do not specify the authorization boundary of an individual subtask. A successful trajectory may involve accessing several files even though only one is necessary: execution traces reveal what a subagent accessed, but not necessarily what it needed. This creates a fundamental supervision gap: agents are expected to make specific authorization decisions, while their training data rarely specifies the resource requirements of each delegated subtask.

We observe that Agent Skills provide a natural source of such supervision. Skills encode procedural knowledge describing not only what should be accomplished, but also the sequence of operations, tools, and workspace resources required at different stages of execution \citep{li2026skillsbench,xu2026agentskills,yang2026skillsurvey}. When a particular step is delegated to a subagent, these procedural requirements provide an explicit basis for determining which resources the subagent should be allowed to access. In this sense, Skills provide information about what a subagent should have access to, rather than merely recording what it happened to access during execution.

Building on this observation, we propose a Skill-driven data synthesis method that converts procedural knowledge into executable supervision. Given a Skill, we synthesize a concrete task and step-by-step procedure, execute the procedure through subagent delegation in an interactive environment, and verify the resulting trajectories. The resulting data is then used for supervised fine-tuning (SFT), enabling agents to learn task specific authorization alongside task execution.

We evaluate our method on ClawArena-Team \citep{xiong2026clawarenateam} for subagent management and resource allocation, and additionally on Terminal-Bench 2.1 \citep{laude2026terminalbench} and SkillsBench \citep{li2026skillsbench} to evaluate task execution performance beyond subagent authorization. On ClawArena-Team, our approach substantially improves tool and workspace permission precision while also improving overall task success. Ablations show that these gains cannot be explained by trajectory quantity alone: removing procedural supervision or replacing verification-based selection with random filtering consistently degrades performance. On Terminal-Bench 2.1 and SkillsBench, our method remains competitive with existing agent data synthesis approaches.

Our contributions are the followings:
\begin{itemize}
    \item We identify \textbf{precise, task-conditioned subagent authorization} as a distinct learning problem in agent orchestration, where the main agent must provide sufficient but non-excessive tools and workspace resources to delegated subagents.
    \item We show that \textbf{Agent Skills provide a natural source of authorization supervision}: their procedural knowledge specifies the operations and resources required at each step, enabling a principled mapping from procedures to authorization ground truth.
    \item We propose a \textbf{Skill-driven data synthesis method} that converts procedural knowledge into verified execution supervision and demonstrate substantial improvements in subagent management on ClawArena-Team, while remaining competitive on Terminal-Bench 2.1 and SkillsBench.
\end{itemize}

\section{Related Work}

\subsection{Agent Training and Subagent Orchestration}

LLM agents have evolved from single-turn tool use toward long-horizon interaction and multi-agent collaboration. ReAct \citep{yao2023react} established a widely adopted paradigm for interleaving reasoning with environment actions, while frameworks such as AutoGen \citep{wu2023autogen}, TDAG \citep{zhu2024tdaACLg}, and OpenHands \citep{wang2024openhands} support multi-agent coordination and tool- and file-based interaction. Recent work further studies how agents should decompose tasks and delegate subtasks, including orchestration benchmarks \citep{liu2026orchestrabench,park2026orchbench,choi2026decisionbench} and methods for learning orchestration behavior such as Chain-of-Agents \citep{li2025chainofagents}, Collator \citep{guo2026lemon}, and MetaAgent-X \citep{chen2026metaagentx}. These works primarily focus on \emph{when} and \emph{to whom} to delegate, leaving the resource scope of delegated subagents largely unspecified.

A parallel line studies the synthesis of training data for tool-using and agentic models. APIGen \citep{liu2024apigen}, ToolACE \citep{liu2024toolace}, ToolMind \citep{yang2025toolmind}, TaskCraft \citep{shi2025taskcraft}, and StateGen \citep{gao2026stategen} generate tool-use or multi-step trajectories through structured synthesis and verification, while TDScaling \citep{chen2026tdscaling} studies the role of trajectory diversity. More closely related to our setting, SearchSwarm \citep{zhang2026searchswarm} trains agents to delegate long-horizon research subtasks, and SKT \citep{sun2026skt} learns skill-use behavior from skill-grounded trajectories. However, these approaches supervise task solving, tool use, or delegation behavior rather than the authorization scope assigned to each delegated subtask. 

\subsection{Agent Skills and Procedural Knowledge}

Agent Skills provide reusable capabilities for LLM agents by packaging natural language instructions with executable resources and reference materials. SkillsBench \citep{li2026skillsbench} characterizes Skills as structured packages of procedural knowledge and demonstrates their utility across diverse tasks. Recent work studies Skills as units of capability management, including skill evolution and composition \citep{zheng2026skillcomposer,tang2026skillmas}, skill coverage analysis \citep{lee2026skillcoverage}, and skill internalization \citep{wang2026skill0}. Other studies examine the dependence of Skill effectiveness on task guidance compatibility \citep{han2026sweskillsbench}.

\subsection{Authorization and Least-Privilege Agent Systems}

The growing autonomy of tool-using agents has motivated work on controlling their permissions. MiniScope \citep{zhu2025miniscope} studies least-privilege enforcement for tool-calling agents, while SkillScope \citep{wu2026skillscope} shows that privilege violations in Agent Skills are inherently task-conditioned. AuthBench \citep{yan2026authbench} further finds that coding agents frequently both omit necessary permissions and unnecessary ones. ClawArena-Team \citep{xiong2026clawarenateam} highlights the same challenge in subagent orchestration, where models tend to over-grant workspace permissions. Other work studies authorization risks introduced by delegation and develops mechanisms such as information flow control and approval-based delegation \citep{rossi2026childinherits,kim2026masdrift,ye2026safeflow,okafor2026approval}.

These studies establish the importance of precise authorization and develop mechanisms for enforcement, auditing, or evaluation.

\section{Method}
\label{sec:method}

\begin{figure*}[t]
\centering
\resizebox{\textwidth}{0.8\height}{\includegraphics[width=\textwidth]{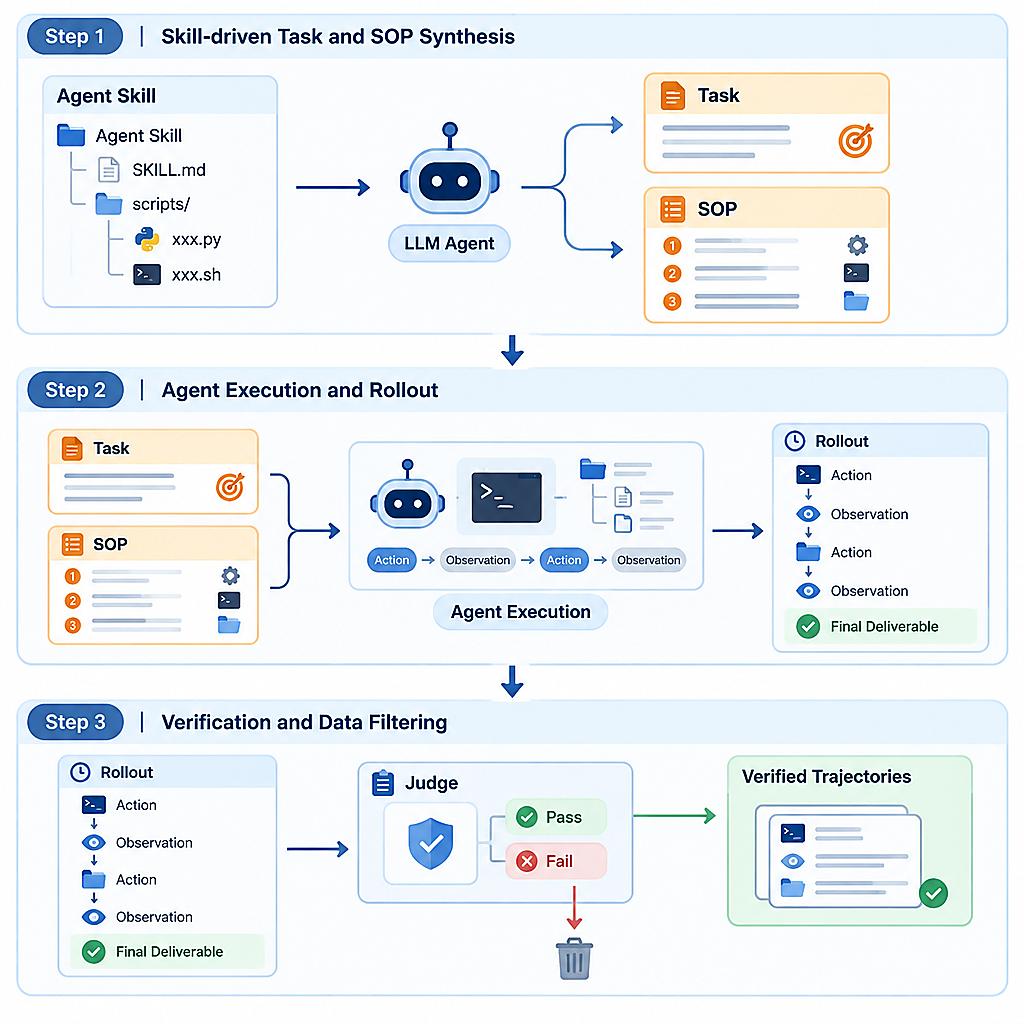}}
\caption{Overview of our Skill-driven data synthesis method.}
\label{fig:method}
\end{figure*}

Our key observation is that Agent Skills encode procedural knowledge that specifies not only what should be accomplished, but also which operations, tools, and workspace resources are required at different stages. We use this structure to construct supervision for task-conditioned subagent authorization: a Skill is first converted into a concrete task and step-by-step procedure (SOP), which guides trajectory synthesis and specifies the resource requirements of delegated steps. The SOP is used only during trajectory synthesis and is removed before SFT training, so the trained agent must infer delegation and authorization decisions from the task itself rather than looking for SOP, as required at benchmark inference time. The overview of our method is shown in Figue~\ref{fig:method}. Unless otherwise specified, the model we use in our method is GLM-5.2~\citep{glm5.2}. The prompts for three stages are shown in Appendix~\ref{app:prompts}, and an example of the generated data is shown in Appendix~\ref{app:example_trajectory}.

\subsection{Problem Setup}
\label{sec:problem}
We consider an agent that solves a task by delegating some subtasks to other agents. For a delegated subtask $s_i$, the main agent provides a subagent instruction $u_i$ together with an authorization scope $a_i=(F_i,T_i)$, where $F_i$ denotes the workspace resources accessible to the subagent and $T_i$ denotes the tools it is allowed to use.

The authorization decision is task-conditioned: the subagent should receive the resources required to complete $s_i$, while access that is unrelated to the delegated subtask should be avoided. However, execution outcomes and trajectories do not directly specify this authorization scope. A successful trajectory may reveal which resources were accessed, but not which of them were necessary for the delegated subtask. Our goal is therefore to construct training supervision that links each delegated step to its required tools and workspace resources.

\subsection{From Agent Skills to Authorization Supervision}
\label{sec:skill_supervision}

The key observation behind our method is that Agent Skills encode procedural knowledge that specifies not only what should be accomplished, but also which operations, tools, and workspace resources are required at different stages of execution. This procedural structure provides a natural source of supervision for task-conditioned subagent authorization.

Given a Skill, we first synthesize a concrete task together with a step-by-step procedure (SOP) describing how the task can be completed. The synthesized task consists of a high-level instruction and a concrete deliverable, while the SOP decomposes the task into a sequence of executable steps $P=(s_1,\ldots,s_n)$. Importantly, the high-level task instruction does not expose these step-level requirements to the agent. Instead, the SOP is used only during trajectory synthesis to make the intended procedure explicit.

For each step $s_i$, the procedure specifies the operation to be performed and the tools or workspace resources needed to perform it. When a step is delegated to a subagent, these requirements provide supervision for constructing its authorization scope. Specifically, the synthesis process associates the delegated step with a subagent instruction and an authorization scope containing the tools and workspace resources required for that step. This establishes a direct correspondence between procedural knowledge and authorization supervision: the procedure specifies \textbf{what needs to be done}, and its resource requirements specify \textbf{what the subagent needs to access}.

To ensure that the synthesized tasks genuinely exercise the capabilities described by the Skill, the task must require the Skill's underlying operations rather than merely mention its functionality. We also make delegation consequential by placing relevant workspace resources behind access control, so that inappropriate authorization decisions can affect whether the delegated step succeeds. These constraints encourage the synthesized data to reflect meaningful relationships between task requirements and authorization decisions rather than arbitrary permission assignments.

The SOP serves only as supervision for trajectory synthesis. It is provided to the teacher agent when generating trajectories so that the teacher can follow the intended procedure and make explicit authorization decisions, but it is removed from the training examples. The trained agent therefore does not learn to rely on the SOP directly; at inference time, it must infer the appropriate delegation and authorization decisions from the task and available environment.

\subsection{Trajectory Synthesis and Verification}
\label{sec:trajectory_synthesis}

\paragraph{Procedure Guided Execution.}
The teacher agent receives the task instruction, the corresponding Skill, and the SOP. The SOP provides synthesis-time guidance for following the intended procedure, deciding which steps to delegate, and determining the resources to authorize for each delegated step. For each delegation, the teacher specifies an authorization scope as an explicit execution constraint for the subagent. The resulting trajectory records the delegation decisions, authorization scopes, tool calls, and workspace interactions.

Access to relevant workspace resources are controlled during execution so that authorization decisions can affect the resulting behavior. This allows the synthesized trajectories to capture the relationship between task execution and authorization decisions rather than treating authorization as an independent annotation.

\paragraph{Independent Verification.}
The generated trajectories are subsequently evaluated by an independent LLM judge. The judge receives the Skill, task instruction, authorization configuration, execution trajectory, and final deliverable, but does not receive the synthesis-time SOP. It evaluates whether the trajectory correctly completes the task, respects the assigned authorization scope, appropriately uses the Skill, and acts naturally as a normal problem solving process. We retain only trajectories that pass all verification criterias.

The SOP is discarded after synthesis and does not appear in the judge procedure or the final training examples. Thus, the resulting data preserves the authorization and execution behavior induced by the Skill while requiring the trained agent to infer delegation and authorization decisions from the task itself rather than the SOP.

\subsection{Validation of the LLM Judge}
\label{sec:judge_validation}

We assess judge reliability by randomly sampling 100 trajectories from the full generated set before filtering. We compare the decisions of the synthesis judge (GLM-5.2) with GPT-6 Sol on the same sample, using identical evaluation inputs and criteria. This comparison measures sensitivity to the choice of judge model. The GLM-5.2 agrees with GPT-6 Sol on 91\% of the overall decisions.

These results support the use of LLM-based verification as a practical filtering mechanism for the synthesized training data. The agreement levels also suggest that the filtering decisions are not highly sensitive to the choice between the two evaluated judge models.

\subsection{Data Statistics}
\label{sec:data_statistics}

\begin{figure}[t]
\centering
\begin{minipage}[c]{0.49\textwidth}
    \centering
    \small
    \setlength{\tabcolsep}{4pt}
    \begin{tabular}{lr}
    \toprule
    Failure Reason & Counts \\
    \midrule
    Incorrect or incomplete deliverable & 895 \\
    Protected-file access violation & 129 \\
    Unnatural or meta-level trajectory & 385 \\
    Skill not genuinely executed & 431 \\
    \bottomrule
    \end{tabular}
    \captionof{table}{Judge failure counts.}
    \label{tab:rejected-dimension-counts}
\end{minipage}\hfill
\begin{minipage}[c]{0.5\textwidth}
    \centering
    \includegraphics[width=\linewidth]{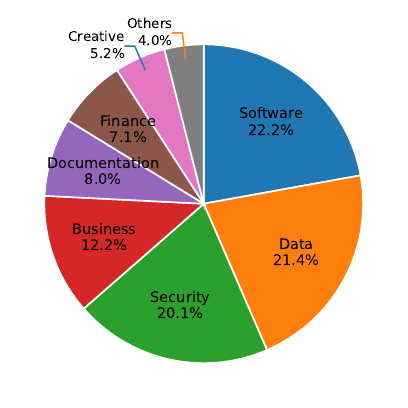}
    \captionof{figure}{Domain distribution of the synthesized dataset.}
    \label{fig:domain_distribution}
\end{minipage}
\end{figure}

We generated 4,413 trajectories in total. Of these,
1,542 trajectories contained at least one type of error, while 2,871
trajectories passed all criteria and were retained as training examples.
Table~\ref{tab:rejected-dimension-counts} reports the failure counts. A single
trajectory may contain multiple errors and therefore contribute to more than
one row in the table; thus, the row counts should not be summed to obtain the
number of erroneous trajectories. Some of the failure cases are shown in Appendix~\ref{app:failure_modes}. The filtering process does not concentrate
the retained data in a single application area: the examples span software
engineering, data analytics, security and compliance, business operations, and
documentation-related tasks. Software engineering, data analytics, and
security and compliance are the three largest domains and together account for
the majority of the dataset. Figure~\ref{fig:domain_distribution} shows the
full domain distribution of the retained examples.

\section{Experiments}
\label{sec:experiments}

\subsection{Experimental Setup}
\label{sec:setup}
We fine-tune the base models using supervised fine-tuning (SFT) on the synthesized training data. The hyperparameters are presented in Appendix~\ref{app:implementation}.

\paragraph{Evaluation benchmarks.}
Our primary evaluation is \textbf{ClawArena-Team} \citep{xiong2026clawarenateam}, a benchmark designed to evaluate subagent management in multi round professional scenarios. In each task, the main agent must create and coordinate subagents, assign each subagent a modality class, a set of tools, and a workspace scope, and complete the delegated work. This setting directly evaluates the ability to make task-conditioned subagent management decisions, making it particularly suitable for evaluating the authorization behavior targeted by our method. We additionally evaluate on \textbf{Terminal-Bench 2.1} \citep{laude2026terminalbench} and \textbf{SkillsBench} \citep{li2026skillsbench} to assess the effectiveness of our method on task execution performance and compare it with existing agent data synthesis approaches.

\paragraph{Metrics.}
For ClawArena-Team, we follow the official evaluation protocol and report its predefined metrics: task completion rate (TCR), tool-permission precision (TPP), read-only compliance (ROC), workspace-permission precision (WPP), modality-choice accuracy (MCA), and the composite Subagent Management Score (SMS). TPP and WPP measure the precision of tool and workspace permissions assigned to subagents, respectively; ROC evaluates compliance with read-only constraints, MCA measures modality-choice accuracy, and TCR indicates overall task success. The benchmark defines SMS as $\mathrm{SMS}=\mathrm{TCR}\times(\mathrm{TPP}+\mathrm{ROC}+\mathrm{WPP}+\mathrm{MCA})/4$. For Terminal-Bench 2.1 and SkillsBench, we report pass rate, with each SkillsBench task evaluated over three trials.

\paragraph{Models and baselines.} We conduct our experiments on two models from the Qwen3.5 \citep{qwen3.5} family: Qwen3.5-4B and Qwen3.5-35B-A3B. We compare our models against the corresponding base model and models trained on publicly released data from Chain-of-Agents (CoA) \citep{li2025chainofagents} and ToolMind \citep{yang2025toolmind}. Chain-of-Agents focuses on learning collaborative problem-solving behavior through multi-agent distillation, while ToolMind provides large-scale tool-use trajectories, including data generated through multi-agent simulation. On Terminal-Bench 2.1 and SkillsBench, we additionally compare against models trained on publicly released data from SWE-smith \citep{swe} and TMAX \citep{tmax}, which target software engineering and terminal-agent tasks. Across methods within each benchmark, we use the same amount of training data.

\subsection{Main Results}
\label{sec:main_results}

\begin{table}[t]
\centering
\begin{minipage}[t]{0.5\textwidth}
\centering
\small
\resizebox{\linewidth}{!}{
\begin{tabular}{lcccccc}
\toprule
Method & SMS & TCR & TPP & ROC & WPP & MCA \\
\midrule
\multicolumn{7}{c}{\textbf{Qwen3.5-35B-A3B}} \\
\midrule
Base     & 30.8 & 45.0 & 59.9 & 96.1 & 35.4 & 82.8 \\
Ours     & \textbf{44.6} & \textbf{56.6} & \textbf{73.3} & \textbf{98.1} & \textbf{59.5} & \textbf{84.5} \\
CoA      & 30.7 & 48.1 & 55.2 & 94.6 & 22.4 & 83.2 \\
ToolMind & 26.2 & 46.5 & 42.3 & 88.7 & 24.1 & 70.2 \\
\midrule
\multicolumn{7}{c}{\textbf{Qwen3.5-4B}} \\
\midrule
Base     & 23.6 & 38.8 & 48.1 & 95.6 & 21.7 & 78.4 \\
Ours     & \textbf{31.9} & \textbf{45.0} & \textbf{57.3} & \textbf{96.7} & \textbf{43.9} & \textbf{85.7} \\
CoA     & 23.7      & 41.1      & 42.4      & 94.5      & 16.3      & 77.6      \\
ToolMind & 19.9 & 39.9 & 30.8 & 89.0 & 13.1 & 66.9 \\
\bottomrule
\end{tabular}
}
\caption{Results on ClawArena-Team (\%).}
\label{tab:main_results}
\end{minipage}\hfill
\begin{minipage}[t]{0.49\textwidth}
\centering
\small
\resizebox{\linewidth}{!}{
\begin{tabular}{lcc}
\toprule
Method & Terminal-Bench 2.1 & SkillsBench (Avg3)\\
\midrule
\multicolumn{3}{c}{\textbf{Qwen3.5-35B-A3B}} \\
\midrule
Base     & 28.1 & 5.7 \\
Ours     & 34.8 & \textbf{10.3} \\
CoA      & 29.2 & \textbf{10.3} \\
ToolMind & 32.6 & 9.2 \\
SWE-smith & 33.7 & 6.9 \\
TMAX     & \textbf{38.2} & 8.0 \\
\bottomrule
\end{tabular}
}
\caption{Results on Terminal-Bench 2.1 and SkillsBench (\%).}
\label{tab:generalization}
\end{minipage}
\end{table}

\paragraph{Results on ClawArena-Team.} As shown in Table~\ref{tab:main_results}, our method achieves the highest
Subagent Management Score (SMS) among the reported results for both
Qwen3.5-35B-A3B and Qwen3.5-4B, outperforming the corresponding base models
by 13.8 and 8.3 points, respectively. Our method also consistently improves
tool-permission precision (TPP) and workspace-permission precision (WPP)
over the base models. In contrast, the Chain-of-Agents and ToolMind
baselines do not show comparable improvements in these authorization
metrics, as their training data provides supervision for agent collaboration
or tool use but does not explicitly supervise task-conditioned authorization
scopes for delegated subagents.

Importantly, the gains in authorization precision do not come at the expense
of task completion. Our method improves the task completion rate (TCR) by
11.6 points for Qwen3.5-35B-A3B and 6.2 points for Qwen3.5-4B over the
corresponding base models. These results show that explicit authorization
supervision can improve permission precision while also improving overall
task performance.

\paragraph{Results on Terminal-Bench 2.1 and SkillsBench.}
Table~\ref{tab:generalization} reports results on Terminal-Bench 2.1 and
SkillsBench. We compare our method with the corresponding base models and
models trained on publicly released data from Chain-of-Agents
\citep{li2025chainofagents}, ToolMind \citep{yang2025toolmind}, SWE-smith
\citep{swe}, and TMAX
\citep{tmax}. Our method achieves competitive performance
against these baselines across the two benchmarks. In particular, while
TMAX is specifically designed for terminal-agent tasks, our method remains
competitive on Terminal-Bench 2.1 and performs strongly on SkillsBench.
These results suggest that training with explicit subagent authorization
supervision remains effective on tasks execution.

\subsection{Ablation Study}
\label{sec:ablation}

\begin{table*}[t]
\centering
\small
\begin{tabular}{lcccccc}
\toprule
Method & SMS & TCR & TPP & ROC & WPP & MCA \\
\midrule
Ours & 44.6 & 56.6 &73.3 &98.1 &59.5 &84.5 \\
Base & 30.8 & 45.0 & 59.9 & 96.1 & 35.4 & 82.8 \\
w/o procedure & 37.5 & 53.5 & 63.6 & 94.7 & 38.3 & 84.0 \\
w/o verification & 32.8 & 49.2 & 57.7 & 96.9 & 30.5 & 81.9 \\
w/o authorization supervision &24.5 & 40.7 & 50.9 & 88.8 & 27.0 & 73.7 \\
\bottomrule
\end{tabular}
\caption{Ablation results on ClawArena-Team.}
\label{tab:ablation}
\end{table*}

We conduct ablation experiments on Qwen3.5-35B-A3B on ClawArena-Team, with results shown in Table~\ref{tab:ablation}.

\paragraph{Effect of procedural guidance.}
Withholding the SOP from the teacher during trajectory generation lowers TPP, WPP, and TCR relative to the full method, but all three metrics remain above the base model. Without explicit procedural guidance, the teacher must infer each step's operations and resource requirements from the Skill documentation and task, increasing the reasoning burden during execution and authorization. The Skill still provides procedural knowledge, so the resulting trajectories retain useful supervision. However, the smaller gains over the base model suggest that explicit SOP guidance helps translate this knowledge into more reliable step-level authorization supervision.

\paragraph{Effect of trajectory verification.}
Removing trajectory verification reduces both tool- and workspace-permission precision below the base model, while TCR remains above the base model but falls short of the full method. Without verification, the training set may retain failed trajectories with incorrect deliverables, unnecessary permissions, or inconsistent delegation behavior. Such examples can introduce conflicting supervision, making it harder for the model to learn reliable relationships between delegated subtasks and their required tools and workspace resources. The results highlight the importance of filtering trajectories for quality rather than relying on execution alone.

\paragraph{Effect of explicit authorization supervision.}
Removing explicit authorization supervision lowers TPP, WPP and TCR relative to both the full method and the base model. In this variant, authorization information is stripped from subagent creation calls in the training trajectories while the remaining execution traces are retained. This does not merely remove a useful learning signal: it presents delegation without the required authorization scopes as a demonstration to imitate, actively reinforcing incorrect delegation behavior. The resulting degradation in permission precision, together with reduced task completion, is consistent with inappropriate resource assignment impairing delegated execution. These results underscore that correct authorization supervision is essential to learning both precise resource assignment and effective task execution.

\subsection{Case Study}
\label{sec:case_study}

To better understand how the learned authorization behavior differs from
that of the base model, we present a representative case study from
ClawArena-Team. The task asks the main agent to review satellite imagery
and supporting survey and legal materials to determine whether a reported
boundary change is justified. The task requires multiple specialized
subtasks, including image analysis, data analysis, and legal
interpretation. In the concrete episode, the main agent delegates three
subtasks: an image analyst reads the two satellite scenes and their
metadata, a survey analyst checks the GNSS measurements and GIS standards,
and a legal analyst reads the boundary definition. The subagents are
intended to return evidence to the main agent, which must cross-reference
the three sources and decide whether the reported change is substantiated.

\begin{figure*}[t]
    \centering
    \includegraphics[width=\textwidth]{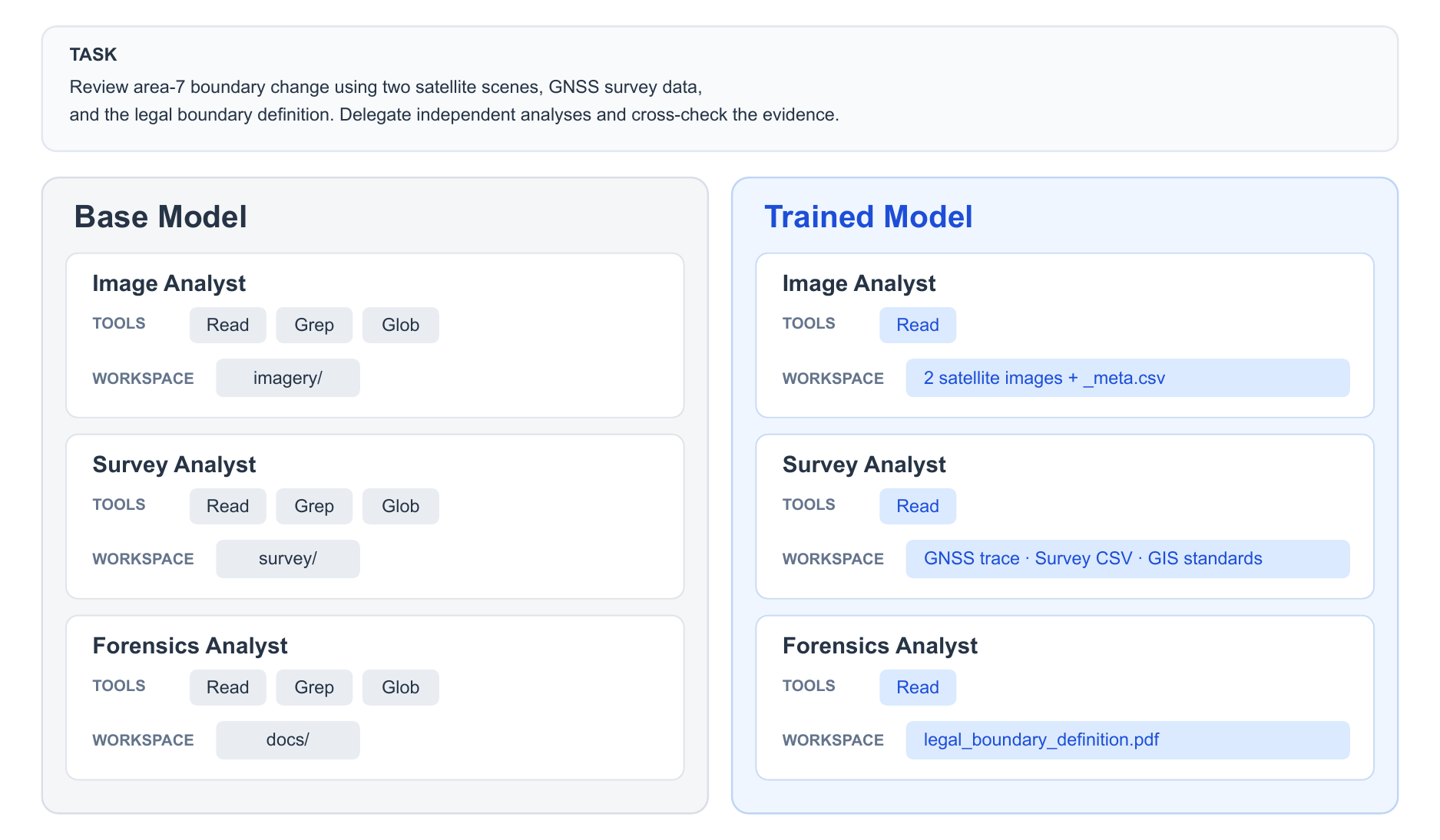}
    \caption{Comparison of subagent authorization by the base and trained models on a satellite boundary-change task. The trained model assigns task-specific tool and workspace scopes while maintaining successful task completion.}
    \label{fig:case_study}
\end{figure*}

Figure~\ref{fig:case_study} illustrates a clear difference in authorization
behavior. For the image analysis subtask, the base model grants access to
the entire \texttt{imagery/} directory with \texttt{Read}, \texttt{Grep},
and \texttt{Glob}, whereas our model grants access only to the two scene
files and \texttt{imagery/\_meta.csv}, with \texttt{Read}. The survey
analyst receives the entire \texttt{survey/} directory and the same three
tools from the base model, but our model grants only the GNSS trace, survey
CSV, and GIS standards file, with \texttt{Read}. The legal analyst is given
the entire \texttt{docs/} directory and \texttt{Read}, \texttt{Grep}, and
\texttt{Glob} by the base model, while our model grants only
\texttt{docs/legal\_boundary\_definition.pdf} and \texttt{Read}. Thus, the
trained model substantially narrows both workspace and tool scopes while
retaining access to the resources required by each delegated subtask.

Importantly, the narrower authorization does not prevent successful task
execution. In the specified runs, the base model completes four of five
rounds, whereas the trained model completes all five rounds. The
corresponding scenario-level TPP and WPP values are $60.0\%$ and $63.3\%$
for the base model, compared with $100.0\%$ and $100.0\%$ for the trained
model. This example illustrates that the learned behavior is not simply to grant
fewer permissions. Instead, the model learns to align authorization scopes
with the requirements of individual delegated subtasks, reducing
unnecessary access while preserving the resources needed for successful
execution.

Additional case studies are provided in Appendix~\ref{app:case_studies}.

\section{Conclusion}
\label{sec:conclusion}

We presented a Skill-driven data synthesis method for learning precise
subagent authorization. By leveraging the procedural knowledge encoded in
Agent Skills, our method constructs verified training trajectories with
explicit authorization supervision. Experiments on ClawArena-Team show
improved tool and workspace permission precision alongside higher task
completion, while results on Terminal-Bench 2.1 and SkillsBench remain
competitive with existing agent data synthesis approaches. These findings
highlight the potential of procedural knowledge as a source of supervision
for agent capability learning.

\addtolength{\bibsep}{-0.5pt}
\bibliography{colm2026_conference}
\bibliographystyle{colm2026_conference}

\newpage
\onecolumn

\appendix
\section{Prompts Used in the Synthesis Pipeline}
\label{app:prompts}

This appendix presents the prompts used in the three synthesis stages. Bracketed
items indicate values inserted at runtime. For task synthesis, selected Skill
files are read and truncated, then inserted inline. The execution and judge prompts receive task-specific
workspace paths and SOP derived from the Skill files. The judge
receives the question, authorization information, trajectory, Skill
documentation, and deliverable, but not the SOP.

\subsection{Task Synthesis}
\label{app:prompt-synthesis}

The first stage reads a Skill and creates workspace inputs, a task
specification and the SOP. The following is the task designer prompt template.

\begin{lstlisting}[style=prompt]
You are a task designer for an agent trajectory synthesis pipeline. Relevant Skill files have been provided inline below. Read them, then:

1. Create the workspace input data files directly on disk under [WORKSPACE_DIR]/.
2. Create ONE JSON file at [WORK_DIR]/spec.json describing the task.

Do NOT output file contents to the terminal. Do NOT create any files other than the input data files under [WORKSPACE_DIR]/ and spec.json.

## Skill content

[SKILL_CONTENT]

## Part 1 -- Task design constraints

### Language

All output must be in English. The Skill may be written in another language, but every produced file and spec field must be in English.

### Skill-centric design

Build the task around the Skill's core functionality. Completing the task must require using the Skill's primary entry points; do not merely inspect its source or reimplement its behavior. At least half of the task steps must use the Skill or process its output. Do not execute the Skill while designing the task.

### Complexity

Create 5-12 coherent steps. Do not add unrelated work just to increase the step count.

### Protected files and delegation

List 0-5 protected files or directories. Roughly 40-50% of tasks should have none. If there are no protected files, all steps must be directly executed by the main agent. If protected files are present, 1-3 steps should be delegated, and every step that needs protected data must be delegated. Use realistic file names that do not explictly reveal that they are protected.

### Input files

Create concrete, realistic, deterministic inputs under [WORKSPACE_DIR]/ and list every path in spec.json. Do not create files elsewhere. Paths in the spec must be relative to the workspace. Do not create or modify output/ or skills/.

### Question and procedure separation

The question is a user-facing deliverable contract, not an execution plan. Keep it under 800 characters. State the exact output path and format, identify key inputs and the Skill in skills/, and give 2-5 observable requirements for the artifact. Do not include steps, commands, script paths, delegation instructions, hidden checks, or the word SOP. Refer to the Skill as "the skill in skills/". Describe protected input as "access-restricted" without explaining how to access it.

### Deliverable

The output must synthesize information produced by executing the Skill. Use json, markdown, text, or files as the deliverable type.

### Delegation precision

For each delegated step, provide a self-contained subagent_prompt, the exact files needed, and the minimal grant_files and grant_tools. grant_tools must be a minimal subset of Read, Write, Edit, Bash, Grep, and Glob.

## Part 2 -- Output format: [WORK_DIR]/spec.json

{
  "task_desc": "<short description>",
  "protected_files": ["<relative paths; 0-5 entries>"],
  "test_input": {"files": ["<relative input paths>"]},
  "question": "<high-level user request>",
  "deliverable": "output/<filename>",
  "deliverable_type": "json|markdown|text|files",
  "steps": [
    {
      "description": "<what the step does and produces>",
      "delegable": false
    },
    {
      "description": "<what the step does and produces>",
      "delegable": true,
      "subagent_prompt": "<task, files, tools, and return format>",
      "grant_files": ["<minimal file set>"],
      "grant_tools": ["<minimal tool set>"]
    }
  ]
}

## How to work

1. Understand the Skill from its inline content and identify its primary entry points.
2. Design a realistic professional task that requires those entry points.
3. Create input files under [WORKSPACE_DIR]/.
4. Decide detailed steps, delegating every step that needs protected data.
5. Write spec.json to [WORK_DIR]/ and list all input files in test_input.files.
\end{lstlisting}

\subsection{Trajectory Execution}
\label{app:prompt-execution}

The second stage runs the task in an access-controlled workspace. The task plan,
workspace location, and protected file list are inserted at runtime.

\begin{lstlisting}[style=prompt]
You are a main agent executing a professional multi-step task in a controlled workspace.

The workspace is at: [WORKSPACE_DIR]
- The complete skill folder is at: [WORKSPACE_DIR]/skills/ (SKILL.md, scripts/, and all resources; run scripts via Bash, set PYTHONPATH/cwd as needed)
- Test input files are already in place.
- output/ is where deliverables go.

## Access control

The following files/directories are PROTECTED -- you CANNOT Read/Grep/Glob them directly. Any step that needs data from these MUST be delegated to a subagent:

[PROTECTED_FILES or "none"]

You CAN directly access skills/, output/, and any other file/directory NOT listed above. If no files are listed as protected, you may read all input files directly -- do not delegate.

## Task details

Here are the operational details for this task -- what needs to be done, in what order, what files each part uses, and which parts require delegation:

[TASK_PLAN]

## Execution rules

1. Work through the task naturally. Do NOT reference any external plan, SOP, or step numbering in your text output.
2. Execute unprotected parts yourself with the appropriate tools.
3. Delegate protected parts with Agent, explicitly stating the authorized files and tools.
4. After producing the deliverable, perform a final delivery audit: check the actual artifact from disk and compare it against every explicit requirement in the user question, including path, format, required sections/keys, evidence, and exclusions. Fix any missing item and check the file before declaring completion.
5. Use only mentioned files and public checks. Do not look for or invoke hidden benchmark verifiers.

Do not output deliverable contents to the terminal. Do not mention SOP, step numbering, DELEGATE, DO-DIRECTLY, or SKILL.md in visible text.
\end{lstlisting}

\subsection{LLM Judge}
\label{app:prompt-judge}

The third stage uses an independent judge session. It receives the task question,
protected file list, delegation grants, trajectory, deliverable, and Skill
documentation, but not the synthesized step procedure.

\begin{lstlisting}[style=prompt]
You are an expert evaluator judging the quality of an AI agent's task execution trajectory.

You have access to files in this workspace. Read the files listed below, then produce a structured evaluation.

## Files to read

1. Skill documentation: [WORKSPACE_DIR]/skills/SKILL.md
2. Task question (provided inline): [TASK_QUESTION]
3. Protected files (provided inline): [PROTECTED_FILES or "none"]
4. Delegation authorization (provided inline): [DELEGATION_GRANTS]
5. Agent trajectory: [WORK_DIR]/trajectory.json
6. Deliverable: [WORKSPACE_DIR]/[DELIVERABLE_PATH]

## Evaluation dimensions

Evaluate each dimension as PASS or FAIL with a one-sentence reason:

1. deliverable_correct: Does the deliverable fulfill the task's core objective with real, substantive data?
2. no_protected_violation: Did the agent respect access control? If there are no protected files and no delegation steps, this is automatically PASS.
3. trajectory_natural: Does the trajectory read like a natural, competent agent? FAIL if you see references to an external plan/SOP/step-numbering.
4. skill_used: Did the agent actually run the Skill's code? Reading its documentation is fine, but the Skill's code must be executed to produce results.

## Output

Write your evaluation as JSON to [WORK_DIR]/judge_verdict.json:

{
  "deliverable_correct": {"pass": <true|false>, "reason": "..."},
  "no_protected_violation": {"pass": <true|false>, "reason": "..."},
  "trajectory_natural": {"pass": <true|false>, "reason": "..."},
  "skill_used": {"pass": <true|false>, "reason": "..."},
  "overall_pass": <true|false>,
  "overall_reason": "one or two sentences summarizing the verdict"
}

overall_pass is true ONLY if all four dimensions pass. Do NOT output anything to the terminal. Just write judge_verdict.json.
\end{lstlisting}

\section{Example of a Verified Synthesized Trajectory}
\label{app:example_trajectory}

To make the synthesized training data concrete, we show a representative
trajectory below. This trajectory contains 27 message entries.
We preserve their complete order and show the key arguments for each tool call;
long system and subagent prompts, source files, tool outputs, and the generated
deliverable are abbreviated for readability.

\begin{lstlisting}[style=prompt]
[00] SYSTEM
You are an autonomous software-engineering agent operating in a sandboxed
workspace. <...>

[01] USER
Build a WeChat conversation playbook for client C-2048 ("Mark") in
output/coaching_playbook.md. Use the warm-talk Skill in skills/ and include
Skill-generated phrase sections, a conversation assessment, and tailored
recommendations. conversation_log.json is access-restricted and must be read
only through a subagent with the required file grant.

[02] ASSISTANT -> Bash
{
  "command": "ls /home/agent/work_mahaodl_warm-talk/work/ && echo \"---SKILLS---\" && ls -R /home/agent/work_mahaodl_warm-talk/work/skills/ | head -50",
  "description": "List workspace and skills directory"
}
[03] TOOL -> Bash
briefing.md ... client_profile.json ... skills/SKILL.md scripts/ ...

[04] ASSISTANT -> Bash
{
  "command": "cd /home/agent/work_mahaodl_warm-talk/work && cat client_profile.json && echo \\\"=====REQ=====\\\" && cat playbook_requirements.json && echo \\\"=====BRIEF=====\\\" && cat coaching_brief.md",
  "description": "Read profile, requirements, brief"
}
[05] TOOL -> Bash
<profile, required sections, coaching brief, and constraints>

[06] ASSISTANT -> Read
{
  "file_path": "/home/agent/work_mahaodl_warm-talk/work/skills/SKILL.md"
}
[07] TOOL -> Read
<warm-talk Skill documentation and usage instructions>

[08] ASSISTANT -> Bash
{
  "command": "cat /home/agent/work_mahaodl_warm-talk/work/skills/_meta.json
              && echo ===PKG=== && cat .../skills/package.json
              && echo ===CHAT.JS=== && cat .../skills/scripts/chat.js",
  "description": "Read skill scripts"
}
[09] TOOL -> Bash
<Skill metadata, package metadata, and chat.js source>

[10] ASSISTANT -> Read
{
  "file_path": "/home/agent/work_mahaodl_warm-talk/work/skills/scripts/chats.js"
}
[11] TOOL -> Read
<phrase-library source defining opening, continue_, response, night,
escalate, safe category lookup, and command-line interface>

[12] ASSISTANT -> Bash
{
  "command": "cd /home/agent/work_mahaodl_warm-talk/work/skills/scripts &&
    echo opening/wechat && node chats.js OPEN_WECHAT &&
    echo opening/dating && node chats.js OPEN_DATING &&
    echo continue/curious && node chats.js CONTINUE_CURIOUS &&
    echo continue/lightFlirt && node chats.js CONTINUE_LIGHT_FLIRT &&
    echo response/sheReplied && node chats.js RESPONSE_SHE_REPLIED &&
    echo response/sheJokes && node chats.js RESPONSE_SHE_JOKES &&
    echo response/sheTeases && node chats.js RESPONSE_SHE_TEASES &&
    echo night && node chats.js NIGHT && echo escalate && node chats.js ESCALATE",
  "description": "Run warm-talk skill for all 9 sections"
}
[13] TOOL -> Bash
<all nine phrase sets; typed entries include tags; continuation entries are
plain strings>

[14] ASSISTANT -> Bash
{
  "command": "node -e ...read chats.js... extract continue_ object...
              console.log(JSON.stringify(continue_, null, 2))",
  "description": "Inspect continue_ data structure"
}
[15] TOOL -> Bash
{"curious":["..."],"lightFlirt":["..."]}

[16] ASSISTANT -> CreateSubagent
{
  "name": "Assess_restricted_chat_log",
  "system_prompt": "Read only conversation_log.json. Return the latest
    unanswered message, sequence number, arc, vibe, one best-fit response
    category, and escalation readiness. Do not write files.",
  "model_key": "llm",
  "tools": ["Read"],
  "accessible_paths": [
    "/home/agent/work_mahaodl_warm-talk/work/conversation_log.json"
  ]
}
[17] TOOL -> CreateSubagent
subagent_id = a24803fdbbde55fb0

[18] ASSISTANT -> RunSubagent
{
  "subagent_id": "a24803fdbbde55fb0",
  "description": "Assess restricted chat log",
  "prompt": "Use the previously specified read-only authorization and return
    the structured conversation assessment."
}
[19] TOOL -> RunSubagent
Latest message: 14 Lina: "You are so annoying" (unanswered).
Vibe: playful-teasing; best-fit category: sheTeases; readiness: ready.

[20] ASSISTANT -> Write
{
  "file_path": "/home/agent/work_mahaodl_warm-talk/work/output/
                coaching_playbook.md",
  "content": "<complete playbook containing profile summary, all nine Skill
              sections, delegated assessment, recommendations, and hold-for-
              later escalation guidance>"
}
[21] TOOL -> Write
File created successfully at output/coaching_playbook.md.

[22] ASSISTANT -> Edit
{
  "file_path": "/home/agent/work_mahaodl_warm-talk/work/output/
                coaching_playbook.md",
  "old_string": "...a very interesting person...",
  "new_string": "...an interesting person...",
  "replace_all": false
}
[23] TOOL -> Edit
File updated successfully.

[24] ASSISTANT -> Read
{
  "file_path": "/home/agent/work_mahaodl_warm-talk/work/output/
                coaching_playbook.md"
}
[25] TOOL -> Read
<complete 6,205-character deliverable; audit confirms all requested sections,
verbatim Skill phrases, delegated assessment, recommendations, and hold flag>

[26] ASSISTANT
The playbook is complete.
\end{lstlisting}

This is a complete interaction trace at the message level rather than a
summary of selected actions. It shows the exact tool sequence and the file paths, command arguments, delegated model, tool grant, and accessible path. The long system prompt, source files, stdout, and final deliverable are shortened only to keep the appendix readable.

\section{Failure Modes in Rejected Trajectories}
\label{app:failure_modes}

The verified set contains only trajectories that pass all four judge criterias,
so the rejected trajectories are useful for diagnosing where synthesis can
break.

\subsection{No Valid Deliverable}

\paragraph{Task.}
The agent must use the Buffer Skill to resolve live connected profiles, apply
campaign rules, schedule approved posts, and write a run report containing the
profiles, scheduled posts, exclusions, and resulting queue.

\begin{lstlisting}[style=prompt,label={lst:failure-live-service}]
WORK ITEM: work_ahmadabugosh_buffer-social
SKILL ACTIONS
  Read: skills/SKILL.md and campaign inputs
  Run: Buffer profile and scheduling commands
ASSISTANT -> Agent
  description: Setup Buffer skill and resolve profiles
  prompt: Read config/buffer_credentials.md, create skills/.env,
  run `node ./buffer.js profiles`, and report the exact result.
TOOL -> Agent
  Async agent launched successfully.
ASSISTANT -> Bash
  command: cd skills && node ./buffer.js profiles
TOOL -> Bash
  HTTP 401 from https://api.buffer.com/graphql
ASSISTANT -> Write
  output/campaign_run.json
TOOL -> Read
  <ERROR>profiles: []; scheduled: []; queue: []</ERROR>
JUDGE
  deliverable_correct: FAIL
  The live profiles were not resolved and no posts were created;
  the report records a blocked run rather than the requested campaign.
\end{lstlisting}

This is a failure to produce the requested deliverable. The HTTP 401 prevented
the agent from accessing live Buffer profiles, so it could neither resolve the
connected profiles nor schedule any posts. Although it wrote a report file,
its profiles, scheduled posts, and queue are all empty: the file records no
completed campaign. Preserving locally excluded rows does not compensate for
the absence of the requested live results. The judge therefore rejects the
deliverable as incorrect, not merely because the service was unavailable, but
because the agent submitted an effectively empty output instead of the
requested campaign run report.

\subsection{Skill Bypass Through Cached or Hand-Written Results}

\paragraph{Task.}
The agent must run the X-news-crawler Skill in hybrid mode for a 72-hour
window, then summarize the structured crawl output and produce a confidence
report.

\begin{lstlisting}[style=prompt,label={lst:failure-skill-bypass}]
WORK ITEM: work_leeguooooo_x-news-crawler
ENVIRONMENT CHECK
  CDP probe: port 9333 is unavailable / occupied by an unrelated process
  Browser session: no usable logged-in Chrome connection
ASSISTANT -> Bash
  command: command -v abs jq python3 pnpm; command -v google-chrome
TOOL -> Bash
  abs: not found; google-chrome: not found
ASSISTANT -> Bash
  command: curl -s http://127.0.0.1:9333/json/version
TOOL -> Bash
  CDP endpoint unavailable; no logged-in Chrome connection
ASSISTANT -> Bash
  command: ls -la .tmp/ && cat .tmp/openclaw-news.json
TOOL -> Bash
  <ERROR>Use pre-existing .tmp/openclaw-news.json</ERROR>
  (the file was not produced by the crawler in this run)
JUDGE
  skill_used: FAIL
  The crawler wrapper was never executed; the report relies on
  a cached artifact rather than fresh Skill output.
\end{lstlisting}

Unlike the preceding case, the central failure here is not an empty
report but a report whose results were not produced by the Skill. With no
usable logged-in browser connection, the agent never ran the X-news-crawler
wrapper. It instead read a pre-existing JSON file and used that cached data
as the basis for its output. Even if the resulting report contains plausible
news items, those items cannot be attributed to a crawler run for the
requested 72-hour window in hybrid mode. The judge therefore rejects the
trajectory for Skill bypass: a populated deliverable is not valid evidence
of Skill use when its underlying data came from a cache rather than from the
Skill's execution.

\subsection{Protected File Access Outside the Delegation Boundary}

\paragraph{Task.}
The agent must validate the paid hashtag-generator Skill, obtain its output for
three campaign topic/platform pairs, and record the billing-compliance result.
The SkillPay configuration is access-restricted and must be handled through an
authorized subagent.

\begin{lstlisting}[style=prompt,label={lst:failure-authorization}]
WORK ITEM: work_yipintangzsp_hashtag-generator-pay
AUTHORIZED PATH
  config/hashtag-generator.json is restricted to the delegated reader
ASSISTANT -> Agent
  description: Read protected SkillPay configuration
  grant_files: [config/hashtag-generator.json]
  grant_tools: [Read]
TOOL -> Agent
  Delegated reader launched.
ASSISTANT -> Bash
  <ERROR>node -e "const c=require('./config/hashtag-generator.json')
  console.log(c.price_per_call, Object.keys(c))"</ERROR>
TOOL -> Bash
  price_per_call=1; skillpay_api_key present; ...
JUDGE
  no_protected_violation: FAIL
  The protected configuration was accessed directly rather than
  exclusively through the authorized subagent channel.
\end{lstlisting}

The key error occurs before the Skill result is interpreted: the main agent
bypasses the authorization mechanism while inspecting configuration metadata.
The fact that only selected fields were printed does not make the access safe;
the policy constrains the access path, not merely the amount of data exposed.
Consequently, even a correct hashtag result and a correctly formatted billing
record would not make this trajectory admissible. The relevant safeguard is an
access check on every filesystem operation, including commands that parse only
metadata or individual fields.

\subsection{Execution-Metadata Leakage into the Visible Trajectory}

\paragraph{Task.}
The agent must route a batch of inbound AI requests through the multi-model
router and produce a consolidated sign-off artifact with routing decisions,
context migration outcomes, and audit statistics.

\begin{lstlisting}[style=prompt,label={lst:failure-metadata-leak}]
WORK ITEM: work_keven0706_multi-model-router
ASSISTANT -> Bash
  command: run the multi-model-router skill and write the sign-off artifact
TOOL -> Bash
  routing and audit outputs returned
ASSISTANT
  <ERROR>The task's step 7 says: add/update policy_flags entries ...</ERROR>
JUDGE
  trajectory_natural: FAIL
  The response explicitly refers to external task step numbering.
\end{lstlisting}

This case is different from an incorrect tool result: the requested artifact
may be substantively useful, but the recorded interaction no longer resembles
an agent solving the user task from the available workspace. The injected
execution plan is synthesis control information and must not appear in
the agent's visible reasoning or final response. The leakage is introduced at the
conversation/interface boundary, after the underlying routing work. The rollout prompt and final
trajectory audit must therefore reject references to plan, SOP, delegation
labels, or injected step numbers before retaining the example.

\section{Implementation and Experimental Details}
\label{app:implementation}

Table~\ref{tab:training_config} reports the settings of training.

\begin{table}[!htbp]
\centering
\small
\begin{tabular}{ll}
\toprule
Configuration & Value \\
\midrule
GPU model & H800 \\
GPU count for training & 8 (Qwen3.5-4B); 24 (Qwen3.5-35B-A3B) \\
Training epochs & 10 \\
Batch size & 256 \\
Loss normalization & per-token loss \\
Data order & shuffled \\
Long-sample handling & truncation enabled \\
Optimizer & Adam \\
Learning rate & $5\times10^{-6}$ \\
Minimum learning rate & $1\times10^{-6}$ \\
Learning-rate schedule & cosine decay \\
Warmup & 10\% of training \\
Weight decay & 0.1 \\
Adam betas & $(0.9, 0.95)$ \\
Distributed optimizer & enabled \\
Activation recomputation & full, uniform, 1 layer \\
Batching & dynamic; 8{,}192 tokens per GPU \\
Attention / hidden dropout & 0.0 / 0.0 \\
Attention backend & FlashAttention \\
\bottomrule
\end{tabular}
\caption{SFT configuration.}
\label{tab:training_config}
\end{table}

\section{Additional Case Studies}
\label{app:case_studies}

This appendix gives trajectory excerpts rather than only prose summaries. Each
excerpt preserves the decision points relevant to our claim: what the main
agent delegates, which tools and resources it grants, and what happens
at the end of the task. The excerpts are shortened only by omitting ordinary
 responses and repeated retry messages; the authorization and
outcome lines are retained. 

\subsection{Tax Reconciliation: Modality-Conditioned Delegation}

\paragraph{Task.}
The main agent must reconcile a cross-border tax filing from a client file,
a voice memo, foreign language receipts, a scanned rental contract, brokerage
records, correspondence, and tax guidance. The task therefore requires text,
audio, and image-capable delegated analysis.

\begin{lstlisting}[style=prompt,label={lst:tax-trace}]
USER REQUEST
Reconcile the cross-border tax filing using the client file, voice memo,
foreign-language receipts, rental-contract photo, brokerage records,
correspondence, and tax guidance. Produce the filing decision.

MAIN AGENT RATIONALE
The inputs have different formats and sizes, so the main agent delegates
large, protected, audio, and image inputs instead of reading them directly.

BASE MODEL
CreateSubagent: audio-listener
  model_key: omni
  tools: [Read]
  accessible_paths: [voice_memos/]
remaining records: additional delegation and workspace inspection

TRAINED MODEL
CreateSubagent: Listen_to_Liu_Wei_voice_memo
  model_key: omni; tools: [Read]
  accessible_paths: [client_voice.wav]
CreateSubagent: Read_German_rental_invoices
  model_key: llm; tools: [Read]
  accessible_paths: [receipt PDFs]
CreateSubagent: Read_rental_contract_photo
  model_key: vlm; tools: [Read]
  accessible_paths: [rental_contract_photo.png]
CreateSubagent: Extract_PFIC_threshold_from_IRS_guidance
  model_key: llm; tools: [Read]
  accessible_paths: [irs_2026_guidance.html]
\end{lstlisting}

The recorded contrast is not a claim that the base model never selected the
right modality: it did assign the voice memo to an \texttt{omni} reader. The
trained trajectory contains more explicit resource-modality pairs across the
heterogeneous inputs, including the voice memo, receipt/contract images, and
textual records. This is the behavior measured by MCA and illustrates why
authorization includes capability choice as well as file scope.

\subsection{Service Migration Review: Partitioning a Workspace}

\paragraph{Task.}
The main agent reviews a migration that splits a payments monolith into
payments, billing, ledger, and reporting services. The delegated readers must
inspect service-specific documentation and migration notes without receiving
unrelated service trees.

\begin{lstlisting}[style=prompt,label={lst:migration-trace}]
USER REQUEST
Review the migration that splits payments, billing, ledger, and reporting from
the monolith. Inspect service documentation and migration notes and report
whether the proposed split is safe.

MAIN AGENT RATIONALE
The service trees are large and contain separate review targets, so the main
agent delegates repository inspection rather than loading every service into
its own context.

BASE MODEL
CreateSubagent: services-reader
  model_key: llm
  tools: [Read, Grep, Glob]
  accessible_paths: [services/]

TRAINED MODEL
CreateSubagent: Explore_payments_and_billing_services
  model_key: llm
  tools: [Bash, Read]
  accessible_paths: [services/billing/README.md,
                      service-specific payments/billing paths]
\end{lstlisting}

Here the relevant trajectory change is the move from a broad services-directory
grant to a service-specific assignment. The trained run records a reader for
payments/billing material with a bounded path and read-oriented tools. The
excerpt therefore supports a more modest claim than a full workspace
partition: it shows targeted resource assignment within a code-review task,
not that every service was independently isolated in this episode.

\subsection{Clinical-Trial Review: Heterogeneous Evidence}

\paragraph{Task.}
The main agent reviews protocol TR-2026-PHII-091 against recruitment,
regulatory, and IRB requirements, including an IRB decision packet and an
audio transcript.

\begin{lstlisting}[style=prompt,label={lst:protocol-trace}]
USER REQUEST
Review protocol TR-2026-PHII-091 against recruitment, regulatory, and IRB
requirements, including the IRB decision packet and its transcript.

MAIN AGENT RATIONALE
The evidence is distributed across protocol, regulatory, and IRB materials;
the main agent delegates document extraction so each reader handles a defined
evidence source.

BASE MODEL
CreateSubagent: protocol-regulatory-reader
  model_key: llm
  tools: [Read, Glob]
  accessible_paths: [protocols/, related regulatory paths]
repeated reader creation while locating evidence

TRAINED MODEL
CreateSubagent: Read_v3_protocol_recruitment_section
  model_key: llm
  tools: [Read]
  accessible_paths: [protocols/v3_zh_en_de.docx]
CreateSubagent: Unzip_IRB_packet_read_decision_PDF_and_audio_transcript
  model_key: llm
  tools: [Bash, Read]
  accessible_paths: [IRB transcript, IRB packet]
\end{lstlisting}

The protocol excerpt shows the same pattern in a document-heavy setting:
separate evidence streams are assigned to separate readers, and the trained
trajectory reaches full completion while retaining high permission and
modality scores. It complements the satellite example without claiming that
all delegation attempts in the run were successful or uniformly minimal.

\end{document}